\documentclass[10pt,twocolumn,letterpaper]{article}

\usepackage[pagenumbers]{iccv} 

\newcommand{\Tref}[1]{Table~\ref{#1}}
\newcommand{\eref}[1]{Eq.~\eqref{#1}}

\newcommand{\fref}[1]{Fig.~\ref{#1}}
\newcommand{\Fref}[1]{Figure~\ref{#1}}

\usepackage{xcolor}
\newcounter{todos}
\AtEndDocument{\ifnum\value{todos}>0 \PackageWarning{TODOS}{There are \arabic{todos} todos left in this paper! Fix them before submitting the paper!} \fi}

\newcommand{\V}[1]{\ensuremath{\mathbf{#1}}}

\makeatletter
\DeclareRobustCommand\onedot{\futurelet\@let@token\@onedot}
\def\@onedot{\ifx\@let@token.\else.\null\fi\xspace}

\makeatother

\newcommand{\rmvp}{RMVP3D~\cite{han2024nersp}\xspace}
\newcommand{\smvp}{SMVP3D~\cite{han2024nersp}\xspace}

\newcommand{\pandora}{PANDORA~\cite{dave2022pandora}\xspace}
\newcommand{\mvas}{MVAS~\cite{cao2023multi}\xspace}
\newcommand{\nero}{NeRO~\cite{liu2023nero}\xspace}

\newcommand{\nersp}{NeRSP~\cite{han2024nersp}\xspace}

\newcommand{\pisr}{PISR~\cite{chen2024pisr}\xspace}
\newcommand{\dr}{3DGS-DR~\cite{ye20243d}\xspace}
\newcommand{\relight}{Re-3DGS~\cite{gao2023relightable}\xspace}
\newcommand{\gs}{Gaussian Surfels~\cite{dai2024high}\xspace}
\newcommand{\polgs}{PolGS~(Ours)\xspace}

\definecolor{iccvblue}{rgb}{0.21,0.49,0.74}
\usepackage[pagebackref,breaklinks,colorlinks,allcolors=iccvblue]{hyperref}

\def\paperID{6937} 
\def\confName{ICCV}
\def\confYear{2025}
\usepackage{pifont}
\usepackage{bm}
\usepackage{multirow}
\usepackage{graphicx}
\usepackage{overpic}
\usepackage{bbding}
\usepackage[accsupp]{axessibility}
\newcommand\blfootnote[1]{%
	\begingroup
	\renewcommand\thefootnote{}\footnote{#1}%
	\addtocounter{footnote}{-1}%
	\endgroup
}

\author{Yufei Han$^1$, Bowen Tie$^1$, Heng Guo$^{1,2*}$, Youwei Lyu$^1$, Si Li$^{1*}$, Boxin Shi$^{3,4}$, Yunpeng Jia$^1$, Zhanyu Ma$^1$\\
\small{$^1$Beijing University of Posts and Telecommunications}~~~~~
\small{$^2$Xiong'an Aerospace Information Research Institute}\\
\small{$^3$State Key Laboratory of Multimedia Information Processing, School of Computer Science, Peking University}\\
\small{$^4$National Engineering Research Center of Visual Technology, School of Computer Science, Peking University}\\
{\tt\small \{hanyufei, tiebowen, guoheng, youweilv, lisi, mazhanyu\}@bupt.edu.cn}\\
{\tt\small shiboxin@pku.edu.cn~~~xibei156@163.com}
}

\begin{document}
\title{PolGS: Polarimetric Gaussian Splatting for Fast Reflective \\Surface Reconstruction}
\twocolumn[{\renewcommand\twocolumn[1][]{#1}
    \maketitle
    \centering
    \vspace{-0.5em}
    \begin{minipage}[b]{\textwidth}
       \begin{overpic}[width=\textwidth]{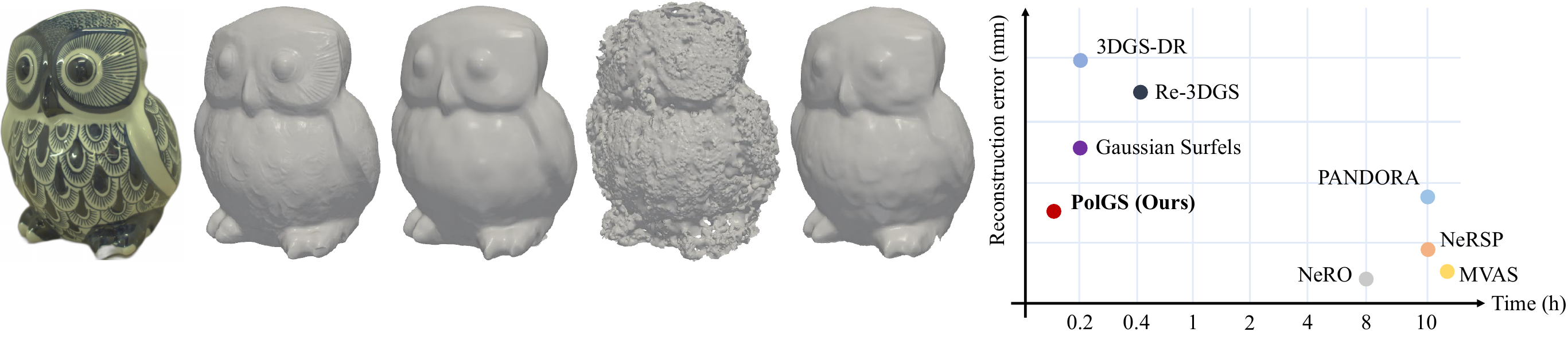}
        \put(3.6,2.7){\color{black}{\fontsize{8.7pt}{1pt}\selectfont Input}}
        \put(14.5,2.7){\color{black}{\fontsize{8.7pt}{1pt}\selectfont \nero}}
        \put(26,2.7){\color{black}{\fontsize{8.7pt}{1pt}\selectfont \pandora}}
        \put(38.5,2.7){\color{black}{\fontsize{8.7pt}{1pt}\selectfont \dr}}
        \put(51.5,2.7){\color{black}{\fontsize{8.7pt}{1pt}\selectfont \polgs}}

        \put(2.35,0.5){\color{black}{\fontsize{8.7pt}{1pt}\selectfont $35$ views}}
        \put(15.5,0.5){\color{black}{\fontsize{8.7pt}{1pt}\selectfont $8$ hours}}
        \put(28,0.5){\color{black}{\fontsize{8.7pt}{1pt}\selectfont $10$ hours}}
        \put(41,0.5){\color{black}{\fontsize{8.7pt}{1pt}\selectfont $12$ min}}
        \put(54,0.5){\color{black}{\fontsize{8.7pt}{1pt}\selectfont $7$ min}}

        \end{overpic}
    \end{minipage}

    \vspace{-0.5 em}
    \captionof{figure}{
		Comparison of efficiency and accuracy on reflective surface reconstruction. 
		Our method takes the shortest time while comparable shape reconstruction accuracy~(measured by Chamfer Distance in millimeters)
		with the existing method based on neural implicit surface representation~\cite{liu2023nero}. 
	}

    \label{fig.teaser}

}]

\begin{abstract}
	Efficient shape reconstruction for surfaces with complex reflectance properties is crucial for real-time
	 virtual reality. While 3D Gaussian Splatting (3DGS)-based methods offer fast novel view rendering
	  by leveraging their explicit surface representation, their reconstruction quality lags behind that of implicit neural representations, particularly in the case of recovering surfaces with complex reflective reflectance.
	    To address these problems, we propose PolGS, a \underline{Pol}arimetric \underline{G}aussian \underline{S}platting 
		model allowing fast reflective surface reconstruction in 10 minutes. By integrating polarimetric constraints into the 
		3DGS framework, PolGS effectively separates specular and diffuse components, enhancing reconstruction quality for 
		challenging reflective materials. Experimental results on the synthetic and real-world dataset validate the effectiveness of our method.
		Project page: \href{https://yu-fei-han.github.io/polgs}{https://yu-fei-han.github.io/polgs}.
\end{abstract}
\vspace{-1em}    
\blfootnote{$^{*}$ Corresponding authors.}
\section{Introduction}

Fast and accurate reconstruction of reflective surfaces is essential for applications like real-time 
virtual reality and inverse rendering. Reflective surfaces meet unique challenges in 3D reconstruction 
as their specular properties require precise handling of light interactions to capture realistic surface 
details. It is desired to develop a fast and accurate reconstruction method for reflective surfaces.

For reflective surface reconstruction, most existing methods rely on implicit neural representations~\cite{mildenhall2020nerf}, 
such as Ref-NeRF~\cite{verbin2022ref} and NeRO~\cite{liu2023nero}, which model complex surface details effectively but suffer from slow 
processing times due to their implicit neural network structure. While the signed distance function (SDF) offers a better geometry representation, the MLP network incurs significant time costs. In contrast, the recent 3D Gaussian Splatting (3DGS)~\cite{kerbl20233d} 
technique offers a promising approach for fast novel view rendering through explicit surface 
representation. However, this representation lacks the same level of detail as implicit methods due to 
limitations in representing fine geometry and surface normals. Additionally, 3DGS-based reconstruction 
methods primarily focus on diffuse surfaces, leaving reflective surfaces under-explored and resulting in lower quality reconstructions for these challenging materials.

To address the challenge of reflective surface reconstruction, polarized images are often utilized to enhance shape representation~\cite{dave2022pandora,han2024nersp,cao2023multi,li2024neisf}. By decomposing the diffuse and specular components according to the polarimetric bidirectional reflectance distribution function (pBRDF) model~\cite{baek2018simultaneous}, the surface normal can be effectively constrained. However, existing polarization-based methods rely on implicit neural representations, which significantly slow down the reconstruction process.

To achieve high efficiency in reflective surface reconstruction, we propose PolGS, a novel method that integrates polarimetric information into the 3DGS architecture for the first time. Unlike previous approaches, the explicit surface representation and splatting  rendering in 3DGS present unique challenges for directly applying polarimetric constraints. To overcome this, we adopt an enhanced 3DGS-based method, \gs, as our baseline, which offers better surface representation capabilities. To ensure that each Gaussian kernel retains a view-independent diffuse color, we modify the spherical harmonics (SH) coefficients to zero-order. 
Next, we introduce a Cubemap Encoder module, inspired by \dr, to extract the specular component. 
Finally, we use the polarimetric constraint during the separation of diffuse and specular components ~\cite{baek2018simultaneous}. This approach resolves shape ambiguities that arise when relying solely on RGB inputs, enabling more accurate reconstruction of \textbf{\emph{reflective}} surfaces, especially for \textbf{\emph{texture-less}} object.

As shown in the teaser~\fref{fig.teaser} and \Tref{table:comparison}, PolGS achieves reconstruction quality comparable with SDF-based methods but 
with over 80-times speed improvement (compared to the fastest and best neural-based reconstruction work in our experiment using RGB input~\cite{liu2023nero}),
 making it highly suitable for virtual reality applications and real-time inverse rendering.

In summary, we advance the reflective surface reconstruction by proposing:
\begin{itemize}
\item PolGS, the first method to incorporate polarimetric information into 3DGS, 
accelerating the progress of surface reconstruction.

\item A pBRDF module integrated into 3DGS, which effectively constrains the diffuse and specular components of reflective surfaces, 
enhancing the accuracy of surface reconstruction.

\item Our method achieves reconstruction quality comparable with SDF-based approaches while significantly 
improving reconstruction time within 10 min. Compared to existing 3DGS methods, PolGS delivers enhancements in 
reconstruction quality.
\end{itemize}

\section{Related works}
Our PolGS aims to reconstruct reflective surface with 3DGS by using polarized images, so we summarize recent progresses in 3D reconstruction techniques, focusing on reflective surfaces through SDF-based methods, 3DGS methods, and polarized image-based reconstruction, respectively.

\vspace{-0.6em}

\paragraph{3D reconstruction based on neural SDF representation}
Novel view synthesis has achieved great success using Neural Radiance Fields (NeRF~\cite{mildenhall2020nerf}). Motivated by the structure of the multi-layer perceptron (MLP) network within NeRFs, numerous 3D reconstruction methods have emerged that leverage implicit neural representations to predict object surfaces. Some approaches~\cite{niemeyer2020differentiable, yariv2020multiview} proposed the signed distance field (SDF) into the neural radiance field, effectively representing the surface as an implicit function. 
Other works~\cite{wang2021neus, yariv2021volume, oechsle2021unisurf, wang2022hf, wang2023neus2,li2023neuralangelo} extend it by proposing efficient framework in detailed surface reconstruction.

These methods based on SDF represent complex scenes implicitly, but they typically suffer from high computational demands and are not suitable for real-time applications. Although some methods~\cite{wang2023neus2, li2023neuralangelo} utilize hash grids and instant-NGP~\cite{mueller2022instant} structure, it is still a challenge for them to reconstruct the mesh efficiently facing the reflective surface. 

Ref-NeRF~\cite{verbin2022ref} uses the Integrated Directional Encoding (IDE) structure to estimate the specular reflection components of the object surface by using predicted roughness, view direction, and surface normals.
NeRO~\cite{liu2023nero} improves it by generating the physically-based rendering (PBR) parameters, and NeP~\cite{wang2024inverse} can better deal with the glossy surface.
TensoSDF~\cite{li2024tensosdf} combines a novel tensorial representation~\cite{chen2022tensorf} with the radiance and reflectance field for robust geometry reconstruction.
However, these approaches can not avoid high optimization time.
\definecolor{newgreen}{rgb}{0.1,0.7,0.1}
\definecolor{3rd}{rgb}{0.95,0.95,0.65}
\definecolor{2nd}{rgb}{1,0.84,0.7}
\definecolor{1st}{rgb}{1,0.7,0.7}
\begin{table}
	\caption{Comparison of different methods in reflective 3D reconstruction: the top four methods are SDF-based, while the bottom four methods are based on 3DGS.}
	\label{table:comparison}
	\small
	\centering
	\resizebox{.5\textwidth}{!}{
		\begin{tabular}{lcccc}
			\toprule
			Input & Method & Reflective & Accuracy & time (h)\\
			\midrule
			RGB Images & \nero & \color{newgreen}\ding{51} &  \colorbox{1st}{high}   &  \colorbox{3rd}{8} \\
			Azimuths & \mvas &  \color{newgreen}\ding{51} & \colorbox{1st}{high}   &  \colorbox{3rd}{11} \\
			Pol. Images& \pandora &  \color{newgreen}\ding{51} & \colorbox{2nd}{medium}  & \colorbox{3rd}{10}\\
			Pol. Images& \nersp &  \color{newgreen}\ding{51} & \colorbox{1st}{high}  & \colorbox{3rd}{10}\\
			\midrule 
			RGB Images & \gs & \color{red}\ding{55} & \colorbox{3rd}{low} &  \colorbox{2nd}{0.2}\\
			RGB Images&  \dr &   \textcolor{newgreen}{\ding{51}} &  \colorbox{3rd}{low} & \colorbox{2nd}{0.2} \\
			RGB Images&  \relight &  \color{newgreen}\ding{51} & \colorbox{3rd}{low} & \colorbox{2nd}{0.4} \\
			Pol. Images&  \polgs &  \color{newgreen}\ding{51} & \colorbox{2nd}{medium} & \colorbox{1st}{0.1} \\
			\bottomrule
			\vspace{-2.8em}	
		\end{tabular}	

	}
\end{table}
\vspace{-1em}	
\paragraph{3D reconstruction based on 3DGS}

3DGS~\cite{kerbl20233d} aims to address limitations of  neural radiance fields by representing complex spatial points using 3D Gaussian ellipsoids. 
However, 3D ellipsoids cannot conform effectively to actual object surfaces, resulting in inaccuracies in shape representation when producing point clouds. 

To overcome these drawbacks, several extensions and modifications have been proposed. SuGaR~\cite{guedon2024sugar} approximates 2D Gaussians with
3D Gaussians, NeuSG~\cite{chen2023neusg}, GSDF~\cite{yu2024gsdf} and 3DGSR~\cite{lyu20243dgsr} integrate an extra SDF network for representing surface normals to supervise the Gaussian Splatting geometry. 
2D Gaussian Splatting~\cite{huang20242d} and Gaussian Surfels~\cite{dai2024high} have taken a different approach by transforming the 3D ellipsoids into 2D ellipses for modeling. This transformation allows for more refined constraints on depth and normal consistency, addressing the surface approximation issues more effectively. 
GOF~\cite{yu2024gaussian} achieves more realistic mesh generation through its innovative opacity rendering strategy.
However, these methods do not focus on reflective surface reconstruction.

\relight associates extra properties, including normal vectors, BRDF parameters,
and incident lighting from various directions to make photo-realistic relighting. 
\dr presents a deferred shading method to effectively render specular reflection with Gaussian splatting. Despite these advancements, these methods still face challenges in many scenarios and cannot provide accurate geometric expression.
\vspace{-1.0em}

\paragraph{3D reconstruction using polarized images}
Polarized images are widely used in Shape from Polarization (SfP)~\cite{miyazaki2003polarization, baek2018simultaneous, smith2018height, deschaintre2021deep, lei2022shape, ba2020deep, lyu2023shape,li2024deep, yang2024gnerp, lyu2024sfpuel}, reflection removal~\cite{li2020reflection,lyu2022physics,wang2025polarized}, and some downstream tasks~\cite{zhou2023polarization, li2023polarized} due to the strong physics-preliminary information in the Stokes field. The SfP task aims to predict the surface normal captured by the polarization camera under the single distant light~\cite{lyu2023shape, smith2018height} or unknown ambient light~\cite{ba2020deep,lei2022shape}. Multi-view 3D reconstruction works using polarized images~\cite{zhao2022polarimetric} try to settle down the $\pi$ and $\pi/2$ ambiguities with the Angle of Polarization (AoP). \pandora first uses polarized images in neural 3D reconstruction work, following the relevant constraints of pBRDF~\cite{baek2018simultaneous}. \mvas leverages multi-view AoP maps to generate tangent spaces for surface points during the optimization process, which can 
reconstruct mesh without rendering supervision. \nersp combines the photometric and geometric cues from polarized images and generates better results under sparse views for reflective surfaces. PISR~\cite{chen2024pisr} focuses on texture-less specular surface 
and integrates the multi-resolution hash grid for efficiency.
NeISFs~\cite{li2024neisf,li2025neisf++} consider the inter-reflection and models multi-bounce polarized light paths during rendering. Despite these advancements, computational cost remains a significant limitation for many polarized-based 3D reconstruction methods.

\section{Preliminaries}
\subsection{Gaussian surfels model}
Our PolGS selects \gs as our base framework due to its stronger geometry expression ability. 
According to \gs, we use a set of unstructured Gaussian kernels $\{G_i=\{\textbf{x}_i,\textbf{s}_i,\textbf{r}_i,o_i,C_i\}|i\in
 \mathcal{N}\}$ to represent the structure of 3DGS, where  $\textbf{x}_i\in\mathbb{R}^3$ denotes the center 
 position of each Gaussian kernel, $\textbf{s}_i=[s^x_i,s^y_i,0]^\top\in\mathbb{R}^3$ is the scaling factors of $x$ and 
 $y$ axes after flatting the 3D Gaussians~\cite{kerbl20233d}, $\textbf{r}_i\in\mathbb{R}^4$ is the rotation quaternion, 
 $o_i\in\mathbb{R}$ is the opacity, and $C_i\in\mathbb{R}^k$ represents the spherical harmonic coefficients of each Gaussian. 
 And Gaussian distribution can be defined by the covariance matrix $\Sigma$ of a 3D Gaussian as:
\begin{equation}
    G(\textbf{x};\textbf{x}_i,\Sigma_i)=\text{exp}(-\frac{1}{2}(\textbf{x}-\textbf{x}_i)^\top\Sigma_i^{-1}(\textbf{x}-\textbf{x}_i)),
\end{equation}
where $\Sigma_i$ can be represented as:
\begin{equation}
\begin{aligned}
	\label{equation:sigma}
    \Sigma_i &= \textbf{R}(\textbf{r}_i)\textbf{s}_i\textbf{s}_i^\top\textbf{R}(\textbf{r}_i)^\top\\
    &= \textbf{R}(\textbf{r}_i)\text{Diag}[(s^x_i)^2,(s^y_i)^2,0]\textbf{R}(\textbf{r}_i)^\top,
\end{aligned}
\end{equation}
where Diag[·] indicates a diagonal matrix and $\textbf{R}(\textbf{r}_i)$ is a $3 \times 3$
rotation matrix represented by $\textbf{r}_i$.
\vspace{-1em}
\paragraph{Gaussian splatting}
According to 3D GS~\cite{kerbl20233d}, novel view rendering process can be represented as: 
\begin{equation}
    C=\sum_{i=0}^nT_i\alpha_ic_i,
\label{eq:gs_render}
\end{equation}
where $T_i=\prod_{j=0}^{i-1}(1-\alpha_j)$ is the transmittance, $\alpha_i = G'(\textbf{u};\textbf{u}_i,\Sigma'_i)o_i$ 
is alpha-blending weight, which is the product of opacity and the Gaussian weight based on pixel $\textbf{u}$. 
In order to speed up the rendering process, the 3D Gaussian in \eref{equation:sigma} 
is re-parameterized in 2D ray space~\cite{zwicker2002ewa} as $G'$:
\begin{equation}
    G'(\textbf{u};\textbf{u}_i,\Sigma'_i)=\text{exp}(-\frac{1}{2}(\textbf{u}-\textbf{u}_i)^\top {\Sigma'_i}^{-1}(\textbf{u}-\textbf{u}_i)),
\end{equation}
where ${\Sigma'_i}= (\textbf{J}_k\textbf{W}_k\Sigma_i\textbf{W}_k^\top\textbf{J}_k^\top)[:2,:2]$. The $\textbf{W}_k$
is a viewing transformation matrix for input image $k$ and $\textbf{J}_k$ is the affine approximation of the projective transformation.
${\Sigma'_i}$ represents the covariance matrix in the 2D ray space.

The depth $\tilde{D}$ and normal $\tilde{N}$ for each pixel can also be
calculated via Gaussian splatting and alpha-blending:
\begin{equation}
        \tilde{D}=\frac{1}{1-T_{n+1}}\sum_{i=0}^{n}T_i\alpha_id_i(\textbf{u}),
\end{equation}
\begin{equation}
        \tilde{N}=\frac{1}{1-T_{n+1}}\sum_{i=0}^{n}T_i\alpha_i\textbf{R}_i[:,2].
\label{eq:gs_normal}
\end{equation}

Specifically, according to~\gs, the depth of pixel $\textbf{u}$ for each Gaussian kernel $i$
is computed by calculating the intersection of the ray cast through pixel $\textbf{u}$ 
with the Gaussian ellipse during splatting. So $d_i(\textbf{u})$ can be represented by Taylor expansion as:
\begin{equation}
    d_i(\textbf{u}) = d_i(\textbf{u}_i) + (\textbf{W}_k\textbf{R}_i)[2,:]\textbf{J}_{pr}^{-1}(\textbf{u}-\textbf{u}_i),
\end{equation}
where $\textbf{J}_{pr}^{-1}$ is the Jacobian inverse mapping one pixel from image space to tangent plane of the Gaussian surfel
as in~\cite{zwicker2001surface}, and $(\textbf{W}_k\textbf{R}_i)$ transforms the rotation matrix of a Gaussian surfel
to the camera space.

\subsection{Polarimetric image formation model}
Polarization cameras can capture polarized images in a single shot, 
which can be represented as $4$ different polarized angles of images 
$I=[I_0,I_{45},I_{90},I_{135}]$. 
The Stokes vector $S = \{s_0,s_1,s_2,s_3\}$ can be calculated by:
\begin{equation}
    S = \begin{bmatrix}\frac{1}{2}
        (I_0+I_{45}+I_{90}+I_{135})\\
        I_{0}-I_{90}\\
        I_{45}-I_{135}\\
        0
    \end{bmatrix},
\end{equation}
where we assume the light source is no circularly polarized here,
thus the $s_3$ is 0.

According to \pandora and \nersp, we assume the incident environmental illumination is unpolarized, 
the Stokes vector 
for the incident light direction $\bm{\omega}$ can be expressed as:
\begin{equation}
    \V{s}_i(\bm{\omega}) = L(\bm{\omega}) [1, 0, 0, 0]^\top,
    \label{eq:env_stok}
\end{equation}
where $L(\bm{\omega})$ represents the light intensity. The polarization 
camera captures outgoing light that becomes partially polarized due to 
reflection, which is modeled using a $4 \times 4$ Muller matrix $\V{H}$. 
The outgoing Stokes vector $\V{s}$ then is formulated as the integral 
of the incident Stokes vector multiplied by this Mueller matrix:
\begin{equation}
    \V{s}(\V{v}) = \int_{\Omega} \V{H} \V{s}_i(\bm{\omega}) \,\, d\bm{\omega},
    \label{eq:out_stok}
\end{equation}
where $\V{v}$ indicates the view direction and $\Omega$ is the integral domain. 
Following the polarized BRDF (pBRDF) model~\cite{baek2018simultaneous}, the 
output Stokes vector can be divided into diffuse and specular components, 
represented by $\V{H}_d$ and $\V{H}_s$ respectively. $\V{s}$ 
can be represented as:
\begin{equation}
    \V{s}(\V{v}) = \int_{\Omega} \V{H}_d \V{s}_i(\bm{\omega}) \,\, d\bm{\omega} + \int_{\Omega} \V{H}_s \V{s}_i(\bm{\omega}) \,\, d\bm{\omega}.
    \label{eq:decomp_stok}
\end{equation}

Based on the derivations from \pandora and \nersp, the output Stokes vector can be 
further specified as:
\begin{eqnarray}
    \V{s}(\V{v}) =   L_d \begin{bmatrix}
        T_o^+ \\
        T_o^- \cos(2 \phi_n) \\
        - T_o^- \sin(2 \phi_n) \\
        0
    \end{bmatrix} +  L_s \begin{bmatrix}
        R^+ \\
        R^- \cos(2 \phi_h) \\
        -R^- \sin(2 \phi_h) \\
        0
    \end{bmatrix},
    \label{eq:photometric_cue}
\end{eqnarray}
where $L_d = \int_{\Omega} \rho L(\bm{\omega})  \bm{\omega}^\top \V{n} \, T_i^+T_i^-   \, d\bm{\omega}$ 
denotes the diffuse radiance associated with the surface normal $\V{n}$, 
Fresnel transmission coefficients~\cite{baek2018simultaneous} $T_{i, o}^+$ and 
$T_{i, o}^-$. The diffuse albedo is represented by $\rho$, and $\phi_n$
is the azimuth angle of the incident light. 

Similarly, $L_s = \int_{\Omega} L(\bm{\omega})  \frac{DG}{4 \V{n}^\top\V{v}}   \,\, d\bm{\omega}$ 
denotes the specular radiance, which involves Fresnel reflection coefficients~\cite{baek2018simultaneous} 
$R^+$ and $R^-$, and the incident azimuth angle $\phi_h$ concerning the half vector 
$\V{h} = \frac{\bm{\omega} + \V{v}}{\|\bm{\omega} + \V{v}\|_2^2}$. For the sake of simplification in our model, we assume that $\phi_h$
is equivalent to $\phi_n$. The Microfacet model incorporates the normal distribution and shadowing terms represented by 
$D$ and $G$~\cite{walter2007microfacet}.

\section{Proposed method}
Our PolGS framework integrates multi-view polarized images, their corresponding masks, and camera pose information to produce a rich output that includes diffuse and specular components represented as Stokes vectors across various views, a reconstructed geometric mesh, and estimated environment light. In this section, we will analyze the surface normal representations employed in SDF-based approaches and 3DGS methods. Then we introduce a theoretical foundation for our polarization-guided reconstruction pipeline.
\vspace{-0.7em}
\begin{figure}
	\centering
	\includegraphics[width=\linewidth]{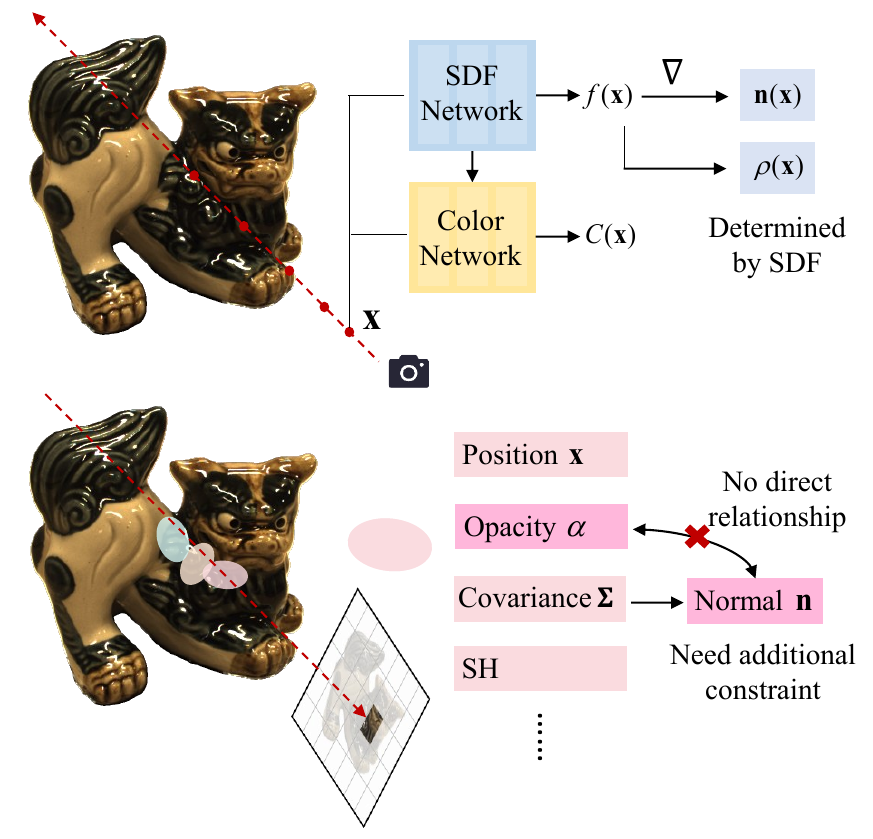}
	
	\caption{Comparison between SDF-based and 3DGS-based method on geometry representation.  (\textbf{Top}) The surface normal of a point has a strong relationship with its opacity in NeuS~\cite{wang2021neus}. 
		(\textbf{Bottom}) The surface normal of a point is dependent on its opacity in \gs.}
	\label{fig:normal-analysis}
    \vspace{-0.5em}
\end{figure}
\subsection{Analysis of surface normal representation}
In implicit neural networks, they use signed distance to determine if points are on the surface 
and the surface normals are obtained by calculating gradients of the SDF. However, 3DGS methods often treat the surface normal as inherent properties typically.
In this section, we analyze the differences
between the surface normal representations in SDF-based methods and 3DGS, explaining 
the reason we use polarimetric cues in the Gaussian Splatting method.
\vspace{-0em}
\begin{figure*}
	\centering
	\includegraphics[width=\linewidth]{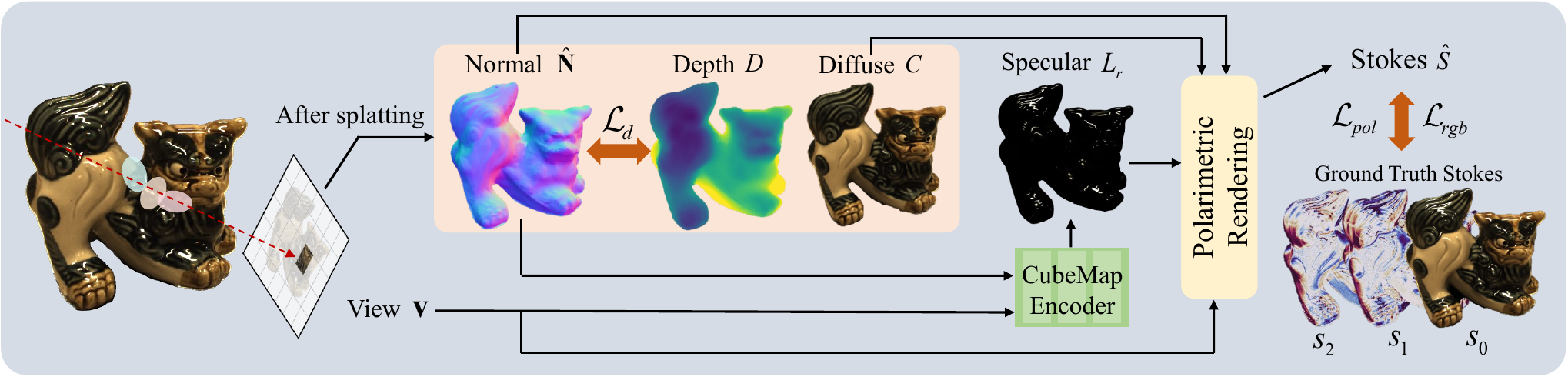}
	\caption{Pipeline of PolGS. We re-rendered Stokes vectors $\hat{s}$ by using the diffuse color $C$ from 3DGS and specular color $L_r$ from Cubemap encoder module, which is supervised by the ground truth Stokes information. }
	\label{fig:pipeline}
	\vspace{-1em}
\end{figure*}
\paragraph{SDF-based representation}
The implicit neural network approach, exemplified by NeuS~\cite{wang2021neus}, enhances the constraints on the reconstructed surface by incorporating a Signed Distance Function (SDF). As illustrated at the top of \fref{fig:normal-analysis}, the SDF value \( f(\textbf{x}) \), generated by the SDF Network, can be used to compute the normal vector \( \textbf{n}(\textbf{x}) \) at point \( \textbf{x} \). According to the definition, the opacity \( \rho(\textbf{x}) \) of a point \( \textbf{x} \)  is given by:

\begin{equation}
    \rho(\textbf{x}) = \max\left(\frac{-\frac{d\Phi_s}{dx}(f(\textbf{x}))}{\Phi_s(f(\textbf{x}))}, 0\right),
\label{eq:neus_density}
\end{equation}
where \( \Phi_s \) is a sigmoid function.

During optimization, the surface normal and the opacity at a given point influence each other. Consequently, after optimization, points farther from the surface tend to have lower opacity values, while the final normal vector is predominantly determined by points located on the surface. This mutual interaction results in a more precise and accurate representation of the surface.

\vspace{-1em}
\paragraph{3DGS-based representation}
In the case of \gs, surface normal is derived from the vector perpendicular to the plane of a 2D ellipsoid. As shown at the bottom of \fref{fig:normal-analysis}, for a single Gaussian kernel, there is no inherent relationship between opacity and the surface normal. 
The pixel-level constraint in 3DGS creates a probabilistic representation where no single Gaussian kernel is definitively assigned to a surface. Consequently, multiple Gaussian kernel configurations can potentially represent equivalent surface geometries, introducing inherent reconstruction ambiguity. 
The optimization mechanism effectively blends contributions from multiple Gaussian points, which offers reconstruction adaptability but compromises the precision of individual point geometric characterizations.

It is obvious that using the normal prediction model as a prior constraint in Gaussian Splatting benefits object reconstruction, as demonstrated by \gs. However, the normal prediction model is not always reliable especially in reflective cases, highlighting the need for a more robust method to provide prior information.
\vspace{-1em}
\paragraph{Polarimetric information for reflective surfaces}
Reflective surface reconstruction is challenging due to the view-dependent appearance. Unlike traditional RGB image inputs, 
polarimetric information can effectively constrain the surface normal during the rendering of Stokes Vectors with the pBDRF model~\cite{baek2018simultaneous}.
 Specifically, as demonstrated in \eref{eq:photometric_cue}, the diffuse and specular components of $s_1$ and $s_2$ 
exhibit a strong correlation with the object's surface normals. Leveraging this property, we incorporate polarimetric information as a prior in shape reconstruction and use the 3DGS-based method to accelerate this process.

\subsection{PolGS}
\paragraph{Network structure}
In this section, we introduce our proposed PolGS, which is a novel 3D reconstruction method that integrates
the Gaussian Splatting method with polarimetric information. The network structure of PolGS is shown in \fref{fig:pipeline}.
The framework comprises two primary components: the Gaussian Surfels module and the Cubemap Encoder module. 
Initially, the Gaussian Surfels module is employed to estimate the diffuse component of the object, 
seems like \gs. Subsequently, we utilize the CubeMap Encoder 
to assess the specular component, akin to the approach taken in \dr. While the CubeMap Encoder does not provide the roughness component, it effectively handles reflective or rough surfaces and maintains high computational efficiency due to its CUDA-based implementation. 
To enhance the rendering process, we incorporate a pBRDF model into the rendering formulation. 
This addition introduces a polarimetric constraint that further refines the Gaussian Splatting method, 
enabling more accurate and realistic 3D reconstructions.
The final rendering formulation model following 
\eref{eq:photometric_cue}
can be represented as:
\begin{equation}
    \hat{S} = C \begin{bmatrix}
        T_o^+ \\
        T_o^- \cos(2 \phi_n) \\
        - T_o^- \sin(2 \phi_n) \\
        0
    \end{bmatrix}+ L_r\begin{bmatrix}
        R^+ \\
        R^- \cos(2 \phi_h) \\
        -R^- \sin(2 \phi_h) \\
        0
    \end{bmatrix},
\label{eq:final_stokes}
\end{equation}
where $C$ is the diffuse color after Gaussian Surfels rendering according to \eref{eq:gs_render} and 
$L_r$ is the specular color after CubeMap rendering $E(\cdot)$, which can be represented as 
$L_r = E(2(\V{v}\cdot \textbf{n})\textbf{n}-\V{v})$.
\vspace{-1em}
\paragraph{Adjustment of spherical harmonic coefficients}
Due to \eref{eq:final_stokes}, the diffuse color rendered by Gaussian Surfels must remain consistent across different viewing directions. To satisfy this constraint, we adjust the spherical harmonic (SH) coefficients of the Gaussian Surfels to zero-order, ensuring the splatting results are view-independent.

\subsection{Training}
The overall training loss in PolGS formulated as a comprehensive weighted sum of multiple loss components:
\begin{equation}
	\mathcal{L} = \mathcal{L}_{rgb}+\lambda_{1}\mathcal{L}_{pol}+\lambda_{2}\mathcal{L}_{m}+\lambda_{3}\mathcal{L}_{o}	+\lambda_{4}\mathcal{L}_{d},
\end{equation}
where we set $\lambda_{1}=1$, $\lambda_{2}=0.1$, $\lambda_{3}=0.01$, 
$\lambda_{4}=0.01+0.1\cdot (iteration/15000)$ to balance the loss function.
\vspace{-1em}
\paragraph{Rendering Stokes loss $\mathcal{L}_{rgb}$ and $\mathcal{L}_{pol}$}
The rendering Stokes loss is combined with the $s_0$ (unpolarized image) rendering loss as 3DGS~\cite{kerbl20233d} and the 
$s_1$, $s_2$ rendering loss as \pandora~\cite{baek2018simultaneous}. These two loss functions can be represented as:
\begin{align}
        &\mathcal{L}_{rgb} = 0.8\cdot L_{1} (s_0,\hat{s}_0)+0.2\cdot L_{DSSIM}(s_0,\hat{s}_0),\\
        &\mathcal{L}_{pol} =L_{1} (s_1,\hat{s}_1)+L_{1} (s_2,\hat{s}_2).
\end{align}
\textbf{Mask loss $\mathcal{L}_{m}$}\quad
The mask loss is used to make the rendering results of the object more accurate, which can be represented as:
\begin{equation}
    \mathcal{L}_{m} = \Sigma\text{BCE}(\textbf{M},\hat{\textbf{M}}).
\end{equation}
\textbf{Opacity loss $\mathcal{L}_{o}$}\quad
The opacity loss follows \gs to encourage the opacity of the Gaussian points to be close to 1 or 0. It can be represented as:
\begin{equation}
    \mathcal{L}_{o} = \Sigma\text{exp}(-{20(o_i-0.5)}^2).
\end{equation}
\textbf{Depth-normal consistency loss $\mathcal{L}_{d}$}\quad
The depth-normal consistency loss follows \gs to make the rendered depth and normal of the object to be more consistent.
It can be represented as:
\begin{equation}
    \mathcal{L}_{d} = 1-\hat{\textbf{N}}\cdot N(V(\hat{\textbf{D}})),
\end{equation}
where $V(\cdot)$ transforms each pixel and its depth to a 3D point and $N(\cdot)$ calculates the normal from neighboring points using the cross product.

\begin{figure*}
	\LARGE
	\begin{overpic}[width=\linewidth]{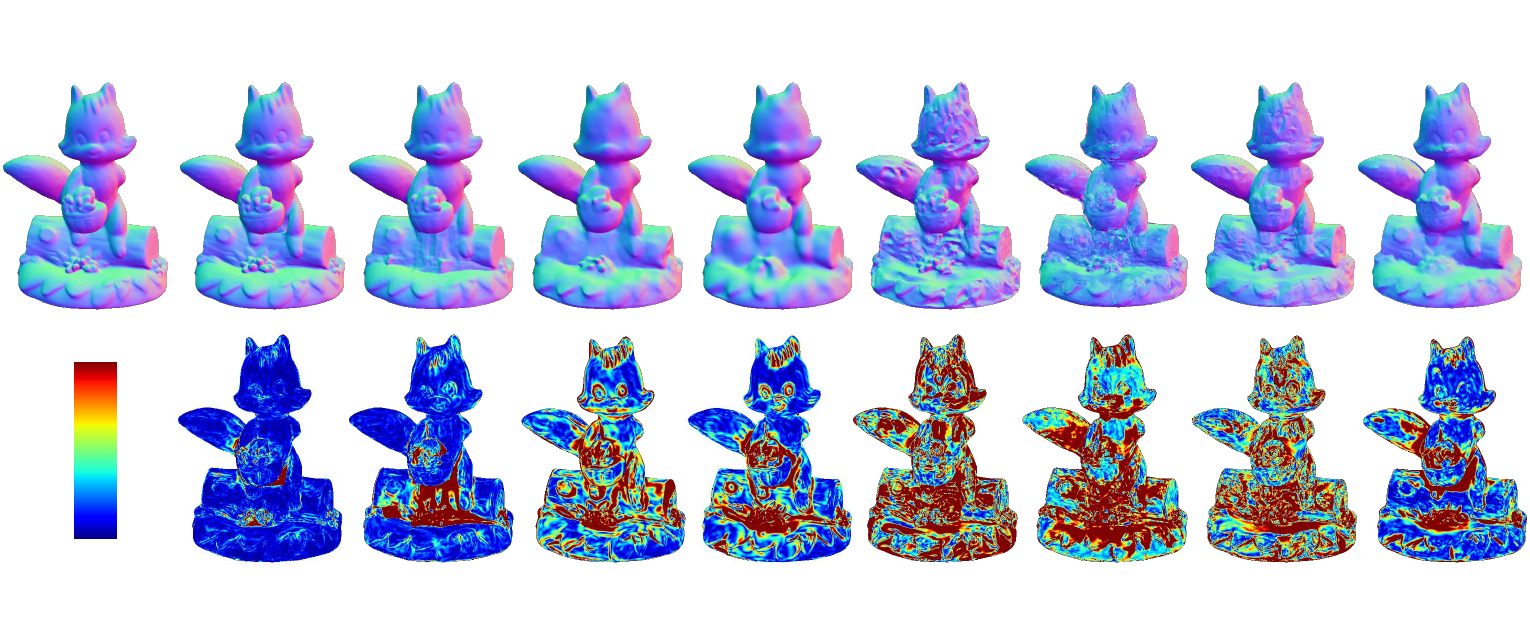}
        \put(5.6,4){\color{black}{\fontsize{8.7pt}{1pt}\selectfont $\text{0}^\circ$}}
        \put(5.3,17.4){\color{black}{\fontsize{8.7pt}{1pt}\selectfont $\text{20}^\circ$}}
        \put(5,36){\color{black}{\fontsize{9pt}{1pt}\selectfont GT}}
        \put(15,36){\color{black}{\fontsize{8pt}{1pt}\selectfont \nero}}
	    \put(26,36){\color{black}{\fontsize{8pt}{1pt}\selectfont \mvas}}        \put(35,36){\color{black}{\fontsize{8pt}{1pt}\selectfont \pandora}}        \put(47,36){\color{black}{\fontsize{8pt}{1pt}\selectfont \nersp}}        \put(55.5,36){\color{black}{\fontsize{8pt}{1pt}\selectfont \gs}}        \put(68.8,36){\color{black}{\fontsize{8pt}{1pt}\selectfont \dr}}        \put(80,36){\color{black}{\fontsize{8pt}{1pt}\selectfont \relight}}        \put(90,36){\color{black}{\fontsize{8pt}{1pt}\selectfont \polgs}}   
	    \put(16,1.7){\color{black}{\fontsize{9pt}{1pt}\selectfont $3.84^\circ$}}
	    \put(27.2,1.7){\color{black}{\fontsize{9pt}{1pt}\selectfont $8.73^\circ$}}
	    \put(38,1.7){\color{black}{\fontsize{9pt}{1pt}\selectfont $11.62^\circ$}}
	    \put(48.7,1.7){\color{black}{\fontsize{9pt}{1pt}\selectfont $10.14^\circ$}}	 
	    \put(59.8,1.7){\color{black}{\fontsize{9pt}{1pt}\selectfont $19.00^\circ$}}	
	    \put(70.8,1.7){\color{black}{\fontsize{9pt}{1pt}\selectfont $19.21^\circ$}}	
	    \put(81.8,1.7){\color{black}{\fontsize{9pt}{1pt}\selectfont $14.53^\circ$}}	
	    \put(92.8,1.7){\color{black}{\fontsize{9pt}{1pt}\selectfont $10.62^\circ$}}	
	    \put(4.4,1.7){\color{black}{\fontsize{8.7pt}{1pt}\selectfont MAE}}

	\end{overpic}
		   \vspace{-1.9em}
	\caption{Qualitative comparisons on surface normal estimation of {\sc Squirrel} in \smvp, where our 3DGS-based method can outperform existing methods based on the same representation and achieves comparable results with SDF-based methods such as \nersp and \pandora.
	}
	\label{fig:syn_normal}
\end{figure*}

\newcommand{\imgw}{0.28}
\begin{figure*}
	\huge
	\resizebox{1\linewidth}{!}{
		\begin{tabular}{@{}c@{}c@{}c@{}c@{}c@{}c@{}c@{}c@{}c@{}}
			Input & \nero & \mvas&	\pandora & \nersp &\gs  &\dr & \relight & \polgs \\  
			\includegraphics[width=\imgw\linewidth]{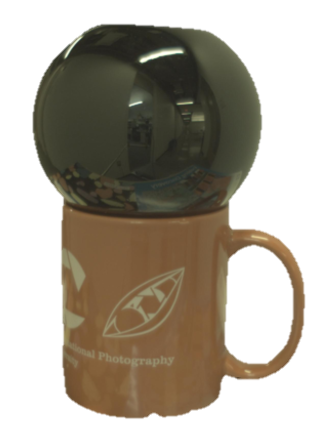}
			&\includegraphics[width=\imgw\linewidth]{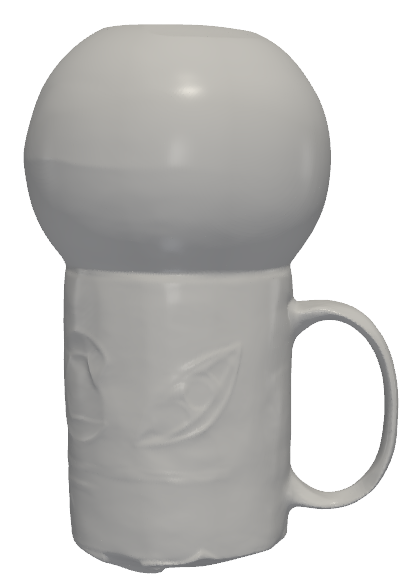}
			&\includegraphics[width=\imgw\linewidth]{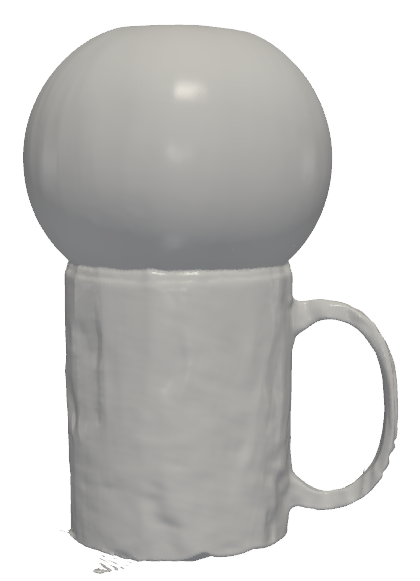}
			&\includegraphics[width=\imgw\linewidth]{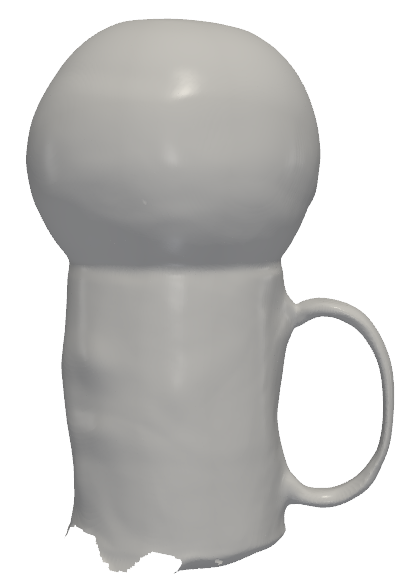}
			&\includegraphics[width=\imgw\linewidth]{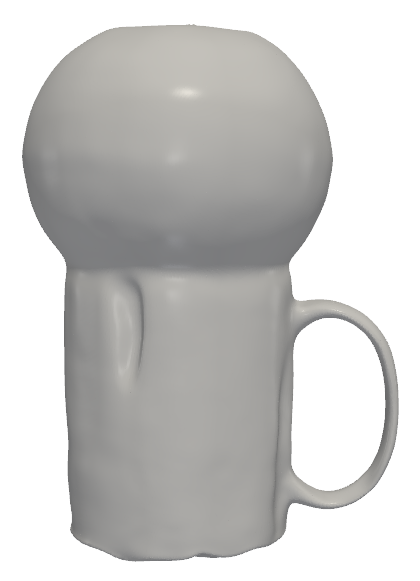}
			&\includegraphics[width=\imgw\linewidth]{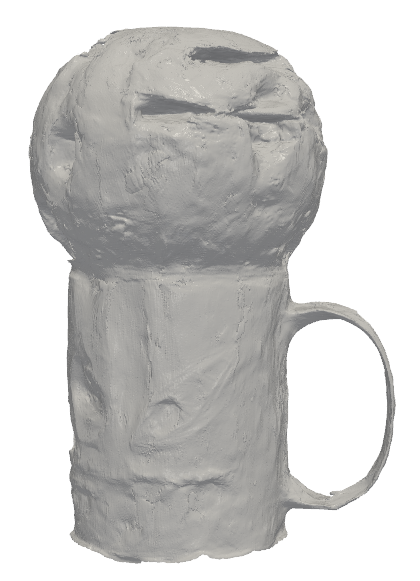}
			&\includegraphics[width=\imgw\linewidth]{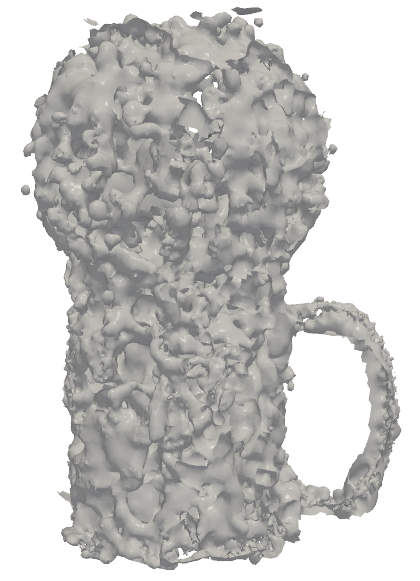}
			&\includegraphics[width=\imgw\linewidth]{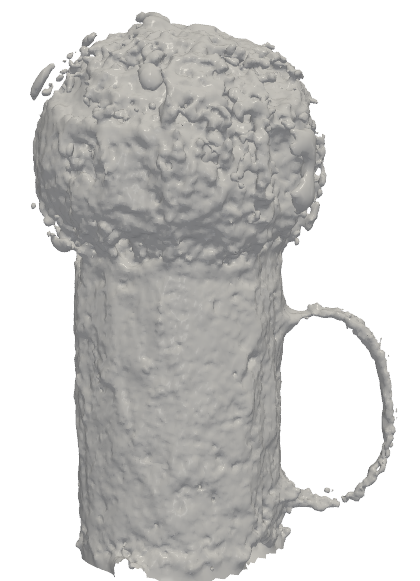}
			&\includegraphics[width=\imgw\linewidth]{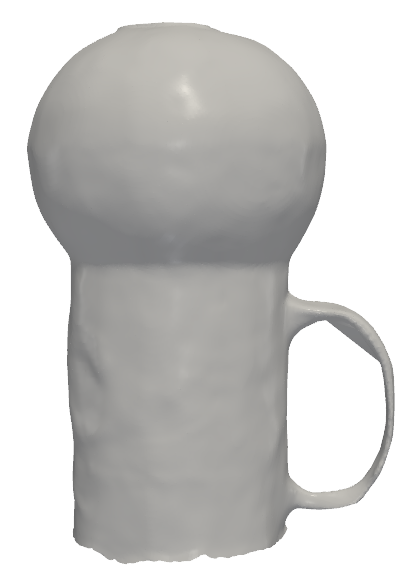}\\
			
			\includegraphics[width=\imgw\linewidth]{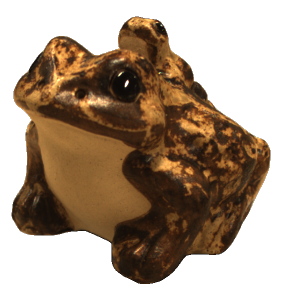}
			&\includegraphics[width=\imgw\linewidth]{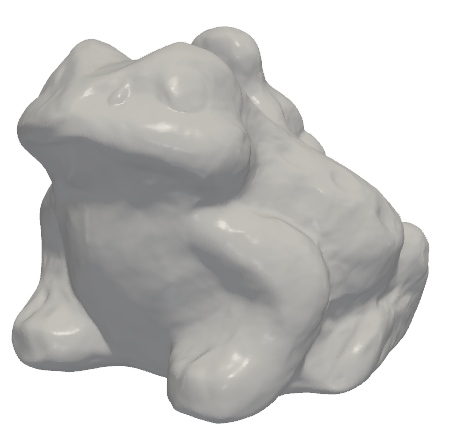}
			&\includegraphics[width=\imgw\linewidth]{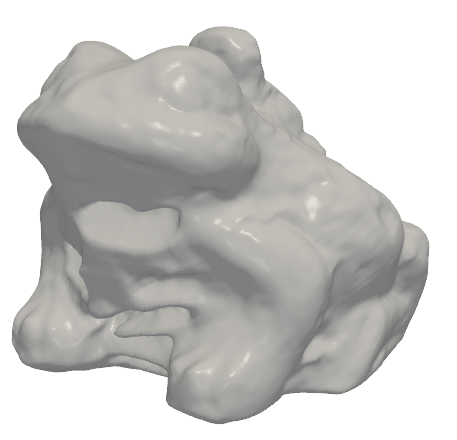}
			&\includegraphics[width=\imgw\linewidth]{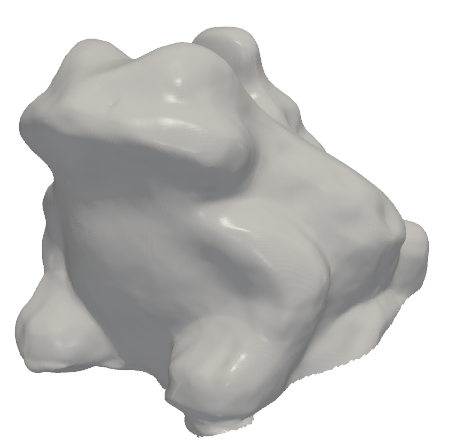}
			&\includegraphics[width=\imgw\linewidth]{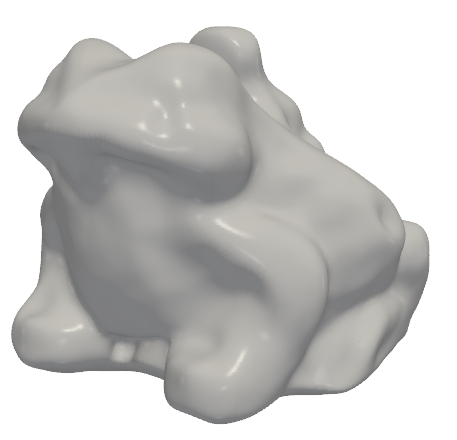}
			
			&\includegraphics[width=\imgw\linewidth]{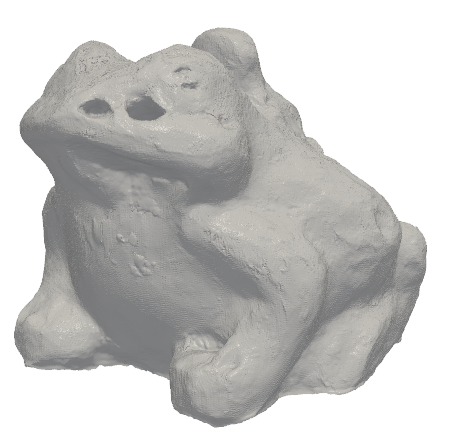}
			&\includegraphics[width=\imgw\linewidth]{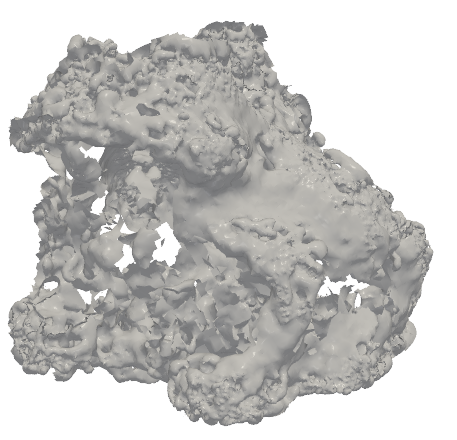}
			&\includegraphics[width=\imgw\linewidth]{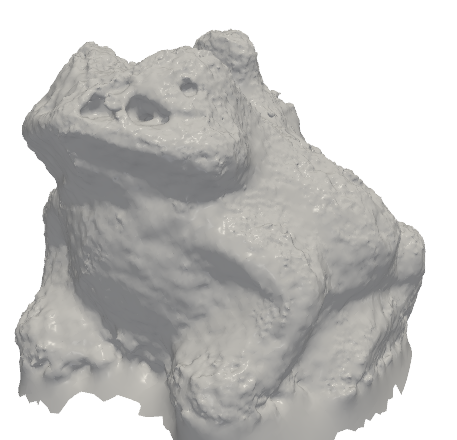}
			&\includegraphics[width=\imgw\linewidth]{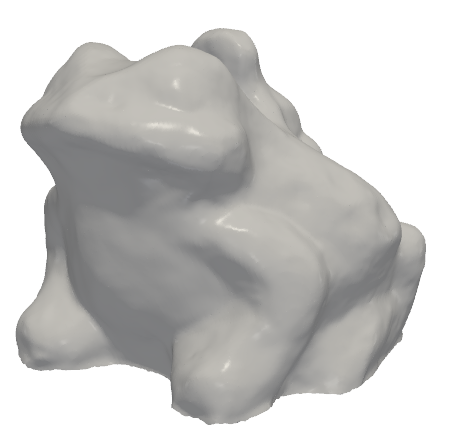}\\
			
		\end{tabular}
	}
	\caption{Qualitative comparison shape on \pandora and \rmvp, where \polgs produces similar quality of reconstruction mesh compared to SDF-based methods. }
	\label{fig:PANDORA-result}
    \vspace{-0.5em}
\end{figure*}
\section{Experiments}
To evaluate the performance of our method, we conduct these experiments: 1) Shape reconstruction 
on synthetic dataset, 2) Shape reconstruction on real-world dataset, 3) Radiance decomposition and 4) Ablation study of polarimetric constraint adding.
\vspace{-1.4em}\paragraph{Dataset}  We use three datasets to evaluate our method: the synthetic dataset \smvp, the real-world dataset \rmvp,
\pandora and \pisr, where \pandora can only be used for qualitative evaluation due to the lack of ground truth.
\vspace{-1.0em}
\paragraph{Baselines} 
We compare our method with state-of-the-art techniques, including SDF-based 
methods \nero, \mvas, \pandora, \nersp, and 3DGS methods \gs, \dr, and \relight. 
All of these methods, except \gs, can handle reflective surfaces. \mvas, \pandora and \nersp utilize 
polarimetric information to reconstruct shapes. Specifically, we do not add the normal prior in \gs among the whole experiments.
\vspace{-1em}
\paragraph{Evaluation metrics}
We use the Chamfer Distance (CD) to evaluate the shape recovery performance and 
the mean angular error (MAE) to evaluate the quality of surface normal estimations. 3DGS-based methods provide point cloud results and we use the Poisson surface reconstruction to generate the mesh for the evaluation.
\vspace{-1.2em}
\paragraph{Implementation details}
We conduct the experiments on an NVIDIA RTX 4090 GPU with 24GB memory. The training of our model is implemented in 
PyTorch 1.12.1 using the Adam optimizer. The specific learning rates of different components
in Gaussian kernels are set the same as \gs. We use the warm-up strategy to train the model, where adding the 
polarimetric information and deferred rendering after 1000 iterations.

\subsection{Reconstruction results on synthetic dataset}

In \fref{fig:syn_normal} and \Tref{table:shape_quantitative_value}, we compare the shape recovery performance of various methods on the \smvp dataset, which contains five objects with spatially varying and reflective properties. However, it is worth noting that SDF-based methods, such as \nero, \mvas, \pandora, and \nersp, outperform 3DGS methods in terms of surface representation. This advantage is largely due to their superior ability to model complex surface details. Among the 3DGS methods, \gs struggles with reflective surfaces, while \dr and \relight still has difficulty in representation of surface normal accurately.
In contrast, PolGS effectively leverages polarimetric information, significantly improving geometric surface performance. Due to the inadequate point cloud generation by other 3DGS methods, they fail to produce reasonable mesh results using Poisson surface reconstruction. Our method, however, achieves the lowest mean Chamfer Distance across the synthetic \smvp dataset, reinforcing the trends observed in surface normal estimations and confirming that our approach delivers the best performance among the 3DGS techniques.

\setlength{\aboverulesep}{-0.7pt}
\setlength{\belowrulesep}{-0pt}
\begin{table*}
	\caption{Comparisons on \smvp and \rmvp evaluated by mean angular error (MAE)~($\downarrow$) with degree and Chamfer distance (CD)~($\downarrow$) in millimeter (mm), respectively. 
	Best and second results in SDF-based and 3DGS-based methods (except PolGS w/o $\mathcal{L}_{pol}$) are highlighted as \colorbox[rgb]{1,0.7,0.7}{1st} and \colorbox[rgb]{1,1,0.4}{2nd}. The time consumed is shown on the far right side of the table.
	}
	\label{table:shape_quantitative_value}
	\vspace{-0.5em}
	\centering
	\LARGE
	\resizebox{\textwidth}{!}{
\begin{tabular}{lcccccccccc|cccccc|cc|c}
	\toprule
	& \multicolumn{10}{c|}{\smvp} & \multicolumn{6}{c|}{\rmvp}&&&
	
	\\
	\cmidrule{2-17}
	 & \multicolumn{2}{c}{\sc Hedgehog } & \multicolumn{2}{c}{\sc Squirrel} & \multicolumn{2}{c}{\sc Snail} & \multicolumn{2}{c}{\sc David} & \multicolumn{2}{c|}{\sc Dragon} & \multicolumn{2}{c}{\sc Shisa} & \multicolumn{2}{c}{\sc Frog} & \multicolumn{2}{c|}{\sc Dog}&\multicolumn{2}{c|}{\multirow{-2}{*}{Mean}}& \\
		\multicolumn{1}{c}{\multirow{-3}{*}{Method}} & MAE & CD & MAE & CD & MAE & CD & MAE & CD & MAE & CD & MAE & CD & MAE & CD & MAE & CD& MAE & CD& \multicolumn{1}{c}{\multirow{-3}{*}{time}}\\
	 \midrule
		 \pandora &9.41&9.50  &10.85& 5.88&8.08&10.97&14.75& 4.88&16.33& 4.78&
	12.93&11.29&15.86&7.88&20.11&10.19&13.54&8.17&10 h\\
	 
	 \nersp& 8.94&6.57 &8.23&  \colorbox[rgb]{1,1,0.4}{3.02}&5.56& \colorbox[rgb]{1,1,0.4}{3.72} &15.38&4.18 &15.30& 3.01&
	 10.79& \colorbox[rgb]{1,1,0.4}{7.39}& \colorbox[rgb]{1,1,0.4}{15.62}& \colorbox[rgb]{1,1,0.4}{6.68}& \colorbox[rgb]{1,0.7,0.7}{16.57}&  \colorbox[rgb]{1,0.7,0.7}{8.57} &12.05&\colorbox[rgb]{1,1,0.4}{5.39}&10 h
	 \\
	
	 \mvas &\colorbox[rgb]{1,1,0.4}{4.30}&\colorbox[rgb]{1,1,0.4}{4.22} & \colorbox[rgb]{1,1,0.4}{6.10}& 3.73 & \colorbox[rgb]{1,1,0.4}{3.30}& 7.87 & \colorbox[rgb]{1,1,0.4}{8.47}& \colorbox[rgb]{1,1,0.4}{3.21} &\colorbox[rgb]{1,0.7,0.7}{8.11}&  \colorbox[rgb]{1,1,0.4}{1.89}&  \colorbox[rgb]{1,1,0.4}{8.56}& 9.28& 17.63& 7.00&  \colorbox[rgb]{1,1,0.4}{16.65}& 8.76&\colorbox[rgb]{1,1,0.4}{9.01}&5.74&11 h\\
	 	
	 \nero &\colorbox[rgb]{1,0.7,0.7}{3.4}&\colorbox[rgb]{1,0.7,0.7}{3.69} & \colorbox[rgb]{1,0.7,0.7}{3.55}& \colorbox[rgb]{1,0.7,0.7}{1.86} & \colorbox[rgb]{1,0.7,0.7}{2.67}& \colorbox[rgb]{1,0.7,0.7}{3.71} &\colorbox[rgb]{1,0.7,0.7}{7.64}& \colorbox[rgb]{1,0.7,0.7}{2.88} &  \colorbox[rgb]{1,1,0.4}{8.12}&\colorbox[rgb]{1,0.7,0.7}{1.69}& \colorbox[rgb]{1,0.7,0.7}{8.41}& \colorbox[rgb]{1,0.7,0.7}{4.88}& \colorbox[rgb]{1,0.7,0.7}{15.29}& \colorbox[rgb]{1,0.7,0.7}{5.39}& 17.72& \colorbox[rgb]{1,1,0.4}{8.74}&\colorbox[rgb]{1,0.7,0.7}{8.27}&\colorbox[rgb]{1,0.7,0.7}{4.10}&8 h\\

    \midrule

    \dr &\colorbox[rgb]{1,1,0.4}{12.28}& 12.66&17.18& 11.20 &11.42& 20.7&20.56&7.91 &26.20 &  9.56  &19.53&15.87&17.08&27.67&24.85&11.00 &19.76&13.32& 0.2 h \\
    \relight &14.40& 19.84&\colorbox[rgb]{1,1,0.4}{15.41}&18.97 &\colorbox[rgb]{1,0.7,0.7}{9.08}& 19.04&\colorbox[rgb]{1,1,0.4}{15.47}& 13.47&\colorbox[rgb]{1,0.7,0.7}{17.65}&  10.96 &
	16.69&14.66&17.85&13.35&21.53 &12.24&\colorbox[rgb]{1,1,0.4}{15.89}&14.12& 0.4 h
	
	 \\
	\gs &16.50&\colorbox[rgb]{1,1,0.4}{8.82}&21.65& \colorbox[rgb]{1,1,0.4}{9.53}&19.05& \colorbox[rgb]{1,1,0.4}{14.04}&21.56& \colorbox[rgb]{1,1,0.4}{7.39}&\colorbox[rgb]{1,1,0.4}{21.41}&  \colorbox[rgb]{1,0.7,0.7}{6.98} &\colorbox[rgb]{1,1,0.4}{12.79}&\colorbox[rgb]{1,1,0.4}{9.09}&\colorbox[rgb]{1,1,0.4}{16.19}&\colorbox[rgb]{1,0.7,0.7}{7.01}&\colorbox[rgb]{1,1,0.4}{19.30}& \colorbox[rgb]{1,1,0.4}{9.56}&18.56&\colorbox[rgb]{1,1,0.4}{10.30}&0.2 h  \\
\polgs &\colorbox[rgb]{1,0.7,0.7}{10.83}&\colorbox[rgb]{1,0.7,0.7}{7.62}&\colorbox[rgb]{1,0.7,0.7}{11.42}& \colorbox[rgb]{1,0.7,0.7}{6.28}&\colorbox[rgb]{1,1,0.4}{9.64}&\colorbox[rgb]{1,0.7,0.7}{10.85}&\colorbox[rgb]{1,0.7,0.7}{13.99}& \colorbox[rgb]{1,0.7,0.7}{5.30}&24.23& \colorbox[rgb]{1,1,0.4}{7.61} &
	\colorbox[rgb]{1,0.7,0.7}{10.88}&\colorbox[rgb]{1,0.7,0.7}{7.76}&\colorbox[rgb]{1,0.7,0.7}{15.03}&\colorbox[rgb]{1,1,0.4}{7.48}&\colorbox[rgb]{1,0.7,0.7}{18.80}&\colorbox[rgb]{1,0.7,0.7}{7.71} &\colorbox[rgb]{1,0.7,0.7}{14.35}&\colorbox[rgb]{1,0.7,0.7}{7.57}&0.1 h\\
    \midrule
    PolGS w/o $\mathcal{L}_{pol}$ & 
    11.39&7.97&11.83&6.39&9.82&10.75&15.44&5.90&25.74&8.22&10.95&7.92&15.16&7.56&18.86&7.74&14.89&7.80&0.1 h\\

    \bottomrule
\end{tabular}	
}
\end{table*}

\subsection{Reconstruction results on real-world dataset}
We further evaluate the reconstruction performance on the \rmvp and \pandora datasets, with qualitative results illustrated in \fref{fig:PANDORA-result} and quantitative results detailed in \Tref{table:shape_quantitative_value}. In real-world scenarios, our method produces reconstructions that align more closely with SDF-based approaches, while significantly outperforming other 3DGS-based methods in terms of reconstruction quality. For example, in the {\sc Vase} case, our approach accurately estimates the shape of a ceramic surface in just $7$ minutes, achieving results closer to those generated by SDF-based methods. 
Additionally, the {\sc Frog} sample highlights our method's ability to reconstruct objects with rough glossy surfaces, showing the robustness and generalization capability of our method. These results collectively demonstrate the effectiveness of our approach in handling diverse real-world objects with varying surface properties.

\begin{figure}
	\begin{overpic}[width=\linewidth, trim={0pt 0pt 0pt 0pt}]{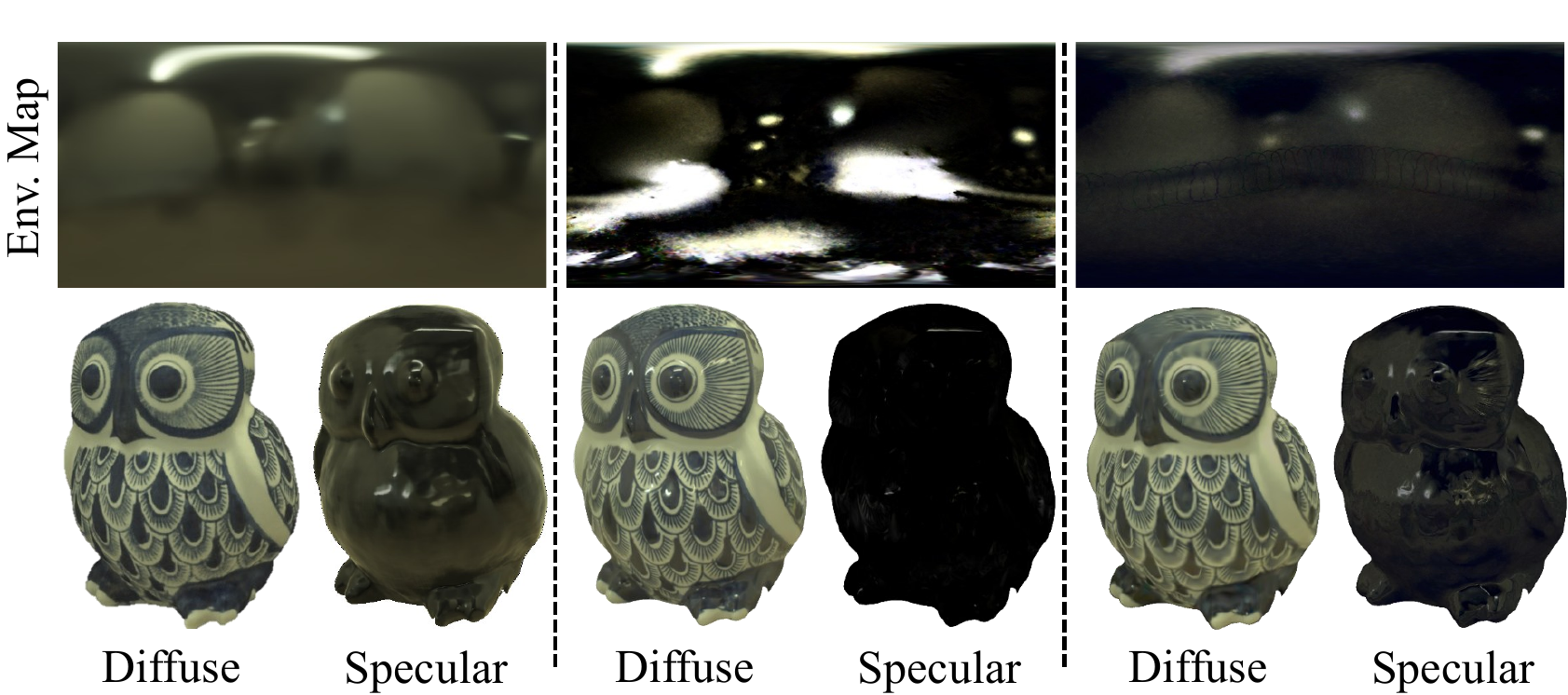}
		
		\put(75,43){\color{black}{\fontsize{8pt}{1pt}\selectfont {\polgs}}}
		\put(10,43){\color{black}{\fontsize{8pt}{1pt}\selectfont {\pandora}}}
		\put(41.5,43){\color{black}{\fontsize{8pt}{1pt}\selectfont {\dr}}}
	\end{overpic}
	\vspace{-1.5em}
	\caption{Separation of diffuse and specular components with \pandora, \dr and \polgs.}
	\label{fig:env_compare}
	\vspace{-2em}
\end{figure}

\subsection{Comparison of radiance decomposition}
\Fref{fig:env_compare} presents a comparison of diffuse and specular component decomposition generated by \pandora, \dr, and \polgs. Here, \pandora utilizes the IDE~\cite{verbin2022ref} structure to produce environment map results, whereas \polgs adopts \dr's methods by employing a Cubemap encoder for the same purpose. 
Compared with \dr, \polgs leverages additional polarimetric information to effectively constrain and disambiguate diffuse and specular components. 
Notably, our results closely align with \pandora's, demonstrating improved radiance decomposition ability.

\begin{figure}
	\centering
\vspace{-1em}
	\begin{overpic}
		[width=\linewidth]{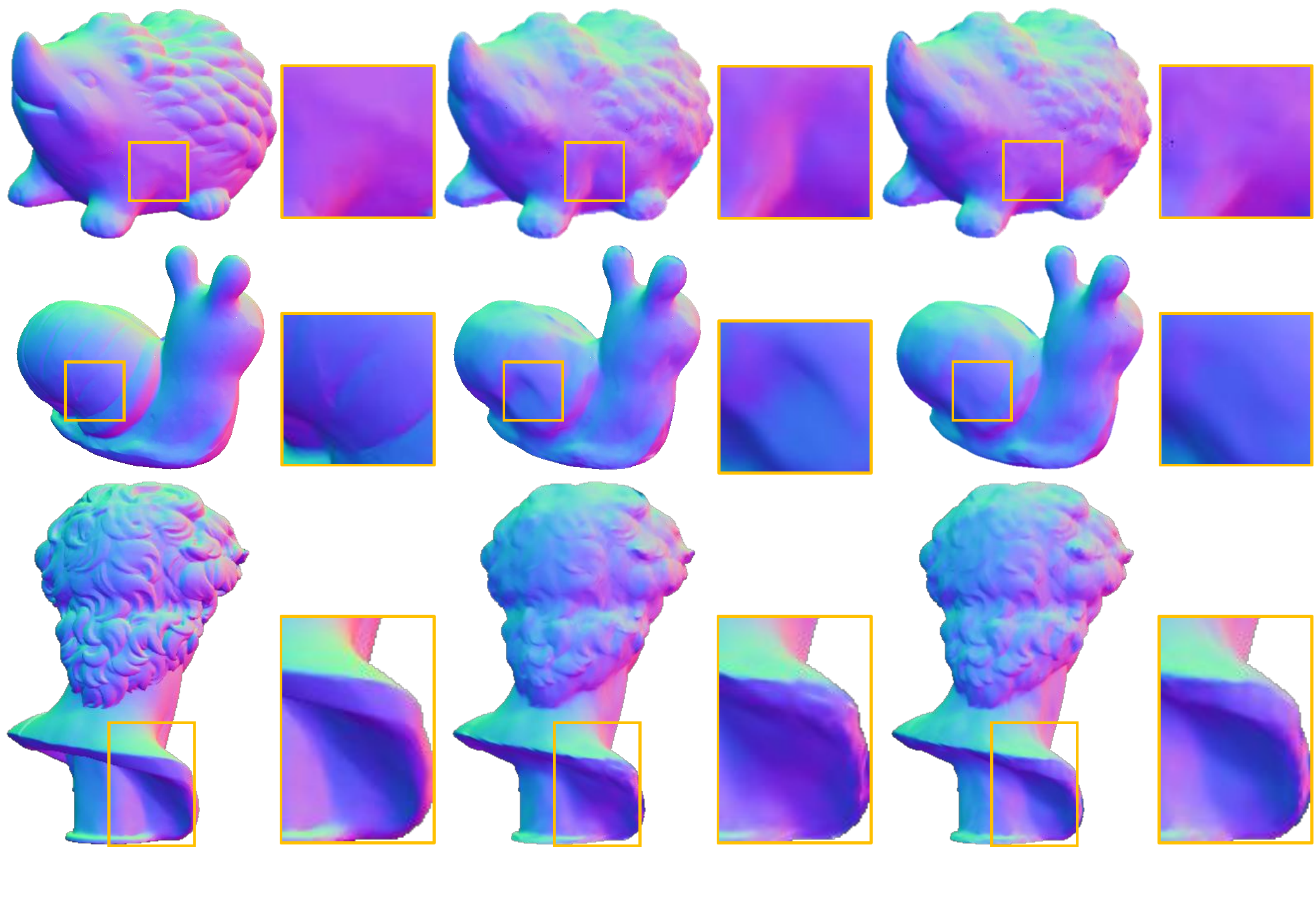}
		\put(12,1){\color{black}{\fontsize{9pt}{1pt}\selectfont GT normal}}
		\put(43,1){\color{black}{\fontsize{9pt}{1pt}\selectfont w/o $\mathcal{L}_{pol}$}}
		\put(80,1){\color{black}{\fontsize{9pt}{1pt}\selectfont w $\mathcal{L}_{pol}$}}
		\put(53,23){\color{black}{\fontsize{8pt}{1pt}\selectfont $14.57^\circ$}}
		\put(87,23){\color{black}{\fontsize{8pt}{1pt}\selectfont $12.51^\circ$}}
		\put(53,65){\color{black}{\fontsize{8pt}{1pt}\selectfont $10.29^\circ$}}
		\put(87,65){\color{black}{\fontsize{8pt}{1pt}\selectfont $9.71^\circ$}}
		\put(53,46){\color{black}{\fontsize{8pt}{1pt}\selectfont $10.24^\circ$}}
		\put(87,46){\color{black}{\fontsize{8pt}{1pt}\selectfont $9.73^\circ$}}
		\put(20,23){\color{black}{\fontsize{8pt}{1pt}\selectfont {\sc David}}}
		\put(20,46){\color{black}{\fontsize{8pt}{1pt}\selectfont {\sc Snail}}}
		\put(19,66){\color{black}{\fontsize{8pt}{1pt}\selectfont {\sc Hedgehog}}}
	\end{overpic}
		\vspace{-1.5em}
	\caption{Ablation study on  surface normal estimation with and without polarimetric information $\mathcal{L}_{pol}$. The MAE values are displayed on the top of each image. }
	\label{fig:ablation}
	\vspace{-1.5em}
\end{figure}

\subsection{Ablation study}\label{sec:ablation}

In this section, we conduct an ablation study to evaluate the contribution of polarimetric information to surface reconstruction. As shown in \Tref{table:shape_quantitative_value} and \fref{fig:ablation}, integrating $\mathcal{L}_{pol}$ consistently improves reconstruction accuracy across both synthetic and real-world datasets. The incorporation of $\mathcal{L}_{pol}$ particularly enhances geometric fidelity in challenging regions such as the {\sc Hedgehog}'s bottom and the {\sc Snail}'s back, where it effectively suppresses concave artifacts.

However, for texture-rich surfaces captured under dense views ($>$30 views), the existing photometric cues (RGB) provide sufficient constraints, leaving limited room for additional improvement from polarimetric information. This contrasts with texture-less surfaces like the {\sc David} statue, where $\mathcal{L}_{pol}$ contributes to reconstruction quality by providing additional geometric constraints.

\begin{figure}
	\vspace{-0.5em}
	\begin{overpic}[width=\linewidth, trim={0pt 0pt 0pt 0pt}]{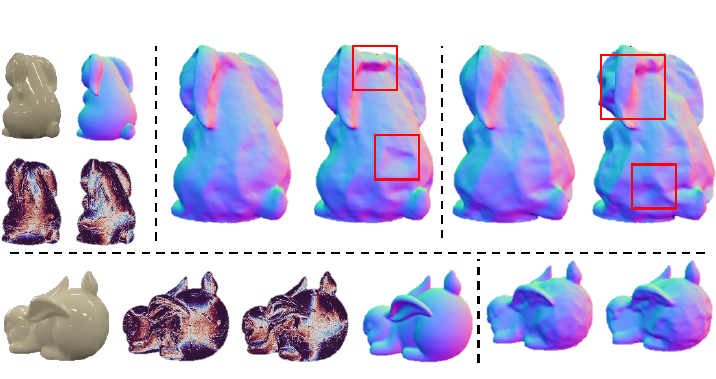}
		\put(17.5,44.5){\color{black}{\fontsize{7pt}{1pt}\selectfont GT}}
		\put(18,30){\color{black}{\fontsize{8pt}{1pt}\selectfont $s_2$}}
		\put(6.7,30){\color{black}{\fontsize{8pt}{1pt}\selectfont $s_1$}}     \put(6.5,44.5){\color{black}{\fontsize{8pt}{1pt}\selectfont $s_0$}}
		
		\put(1,13){\color{black}{\fontsize{8pt}{1pt}\selectfont $s_0$}}
		
		\put(35,13){\color{black}{\fontsize{8pt}{1pt}\selectfont $s_2$}}
		\put(18,13){\color{black}{\fontsize{8pt}{1pt}\selectfont $s_1$}}    
		\put(52,13){\color{black}{\fontsize{7pt}{1pt}\selectfont GT}}
		
		\put(28,19){\color{black}{\fontsize{8pt}{1pt}\selectfont w $\mathcal{L}_{pol}$}}
		\put(47,19){\color{black}{\fontsize{8pt}{1pt}\selectfont w/o $\mathcal{L}_{pol}$}}
		\put(67,19){\color{black}{\fontsize{8pt}{1pt}\selectfont w $\mathcal{L}_{pol}$}}
		\put(85,19){\color{black}{\fontsize{8pt}{1pt}\selectfont w/o $\mathcal{L}_{pol}$}}
		
		\put(35.5,45){\color{black}{\fontsize{7pt}{1pt}\selectfont $13.3^\circ$}}
		\put(55.5,45){\color{black}{\fontsize{7pt}{1pt}\selectfont $15.3^\circ$}}
		\put(75,45){\color{black}{\fontsize{7pt}{1pt}\selectfont $16.5^\circ$}}
		\put(94.8,45){\color{black}{\fontsize{7pt}{1pt}\selectfont $19.0^\circ$}}
		
		\put(37,49){\color{black}{\fontsize{8pt}{1pt}\selectfont {20 views}}}
		\put(75,49){\color{black}{\fontsize{8pt}{1pt}\selectfont {14 views}}}
		
		\put(71,0.5){\color{black}{\fontsize{8pt}{1pt}\selectfont w $\mathcal{L}_{pol}$}}
		\put(86,0.5){\color{black}{\fontsize{8pt}{1pt}\selectfont w/o $\mathcal{L}_{pol}$}}
		\put(85,13.8){\color{black}{\fontsize{7pt}{1pt}\selectfont $13.8^\circ$}}
		\put(67,13.8){\color{black}{\fontsize{7pt}{1pt}\selectfont $11.8^\circ$}}
		
	\end{overpic}
		\vspace{-1.5em}
	\caption{Stokes vectors can help surface normal reconstruction on texture-less surfaces.}
	\label{fig:r1}
	\vspace{-1.5em}
\end{figure}

\paragraph{Boosting on texture-less surfaces recovery with $\mathcal{L}_{pol}$}
To further validate the benefits of $\mathcal{L}_{pol}$ on texture-less surfaces, we conduct additional experiments using a real-world dataset from PISR~[{\color{iccvblue}5}], as shown in \fref{fig:r1}. While the RGB appearance lacks discernible texture, the polarization channels $s_1$ and $s_2$ exhibit meaningful variations that provide additional geometric cues for resolving shape ambiguities. The results show that both concave and convex surface distortions are more accurately reconstructed when $\mathcal{L}_{pol}$ is employed. Moreover, as the number of input views decreases, the advantages of incorporating polarization become even more evident. This demonstrates the unique value of polarimetric information in challenging scenarios where conventional RGB inputs offer limited constraints.

\section{Conclusion}
We propose PolGS, a novel Polarimetric Gaussian Splatting method for fast and accurate 3D reconstruction of reflective surfaces. PolGS leverages polarimetric information for separating the diffuse and specular components based on Stokes field, which can help to constrain the surface normal in the gaussian splatting representation and finally improve the reflective surface reconstruction results.
Extensive experiments demonstrate that PolGS outperforms state-of-the-art 3DGS methods in terms of accuracy and efficiency.

\section*{Acknowledgments}
{
This work was supported by Hebei Natural Science Foundation Project No. 242Q0101Z, Beijing-Tianjin-Hebei Basic Research Funding Program No. F2024502017, National Natural Science Foundation of China (Grant No. 62472044, U24B20155, 62225601, U23B2052, 62136001), Beijing Natural Science Foundation Project No. L242025, BUPT Excellent Ph.D. Students Foundation. We thank openbayes.com for providing computing resource.}

{
    \small
    \bibliographystyle{ieeenat_fullname}
    \bibliography{main}
}

\end{document}